%% file: main.tex
\documentclass[10pt,twocolumn,letterpaper]{article}

\usepackage{cvis}              
\input{preamble}
\definecolor{cvisblue}{rgb}{0.21,0.49,0.74}
\usepackage[pagebackref,breaklinks,colorlinks,allcolors=cvisblue]{hyperref}
\def\confName{CVIS}
\def\confYear{2026}

\title{Beyond Static Gaussians: An Empirical Investigation of Architectural Paradigms for Dynamic 3D Scene Reconstruction}

\author{
Adrian Ramlal\\
University of Waterloo\\
{\tt\small adrian.ramlal@uwaterloo.ca}
\and
John S. Zelek\\
University of Waterloo\\
{\tt\small jzelek@uwaterloo.ca}
}

\begin{document}
\maketitle

\input{sec/0_abstract}

\input{sec/1_introduction}
\input{sec/2_related_work}

\input{sec/3_methods}
\input{sec/4_results}

\input{sec/5_discussion}

{
    \small
    \bibliographystyle{ieeenat_fullname}
    \bibliography{main, references}
}

\end{document}

%% file: sec/0_abstract.tex

\begin{abstract}
Dynamic scene reconstruction via 3D Gaussian Splatting (3DGS) has emerged as a compelling approach for representing evolving environments, yet understanding trade-offs between methodologies remains crucial. This paper presents a comprehensive analysis of dynamic 3DGS methods, categorizing them into two paradigms: structure-guided methods employing auxiliary representations (deformation fields, canonical spaces, grids) to model temporal changes, and gaussian-centric methods encoding dynamics directly into primitives via continuous functions or 4D representations. We evaluate representative methods from both paradigms on the D-NeRF benchmark. Our findings reveal that structure-guided methods achieve superior reconstruction fidelity and compact model sizes, while gaussian-centric approaches demonstrate significantly higher rendering speeds enabling real-time performance, though with greater quality variability and potentially substantial storage overhead. This analysis highlights a fundamental trade-off between reconstruction quality/compactness versus rendering speed, providing insights to guide future research and application development in dynamic scene reconstruction.
\end{abstract}

%% file: sec/1_introduction.tex
\section{Introduction}
\label{sec:introduction}

Dynamic scene reconstruction captures and digitally represents environments where geometry and appearance evolve over time \cite{park2021nerfies}, underpinning applications in VR/AR, interactive media, robotics, and autonomous navigation. However, the task is inherently sparse and under-constrained, relying on limited viewpoints and time samples that leave gaps in data \cite{Cao2023HEXPLANE}. The complexity of modeling non-rigid deformations, abrupt changes, and lighting variations further complicates reconstruction.

3D Gaussian Splatting (3DGS) \cite{kerbl3Dgaussians} has emerged as a powerful solution, offering explicit scene representation and efficient real-time rendering. Methods for adapting 3DGS to dynamic scenes have coalesced into two paradigms \cite{chen_survey_2024}: \textit{structure-guided} methods leverage auxiliary structures, MLP-based deformation fields \cite{yang2023deformable3dgs} or sparse control points \cite{yang2023gs4d}, to explicitly model temporal transformations; \textit{gaussian-centric} methods encode dynamics directly into primitives via continuous functions \cite{lin2024gaussian, katsumata_compact_2024}, Radial Basis Functions \cite{li2023spacetime}, or 4D space-time representations \cite{yang2023gs4d, duan:2024:4drotorgs}.

Our comparative analysis reveals distinct performance characteristics: structure-guided methods achieve higher reconstruction fidelity and temporal consistency with compact models, particularly on monocular benchmarks like D-NeRF \cite{pumarola2020d}, while gaussian-centric methods demonstrate superior computational efficiency and rendering speed essential for real-time interaction, though with increased storage demands. This paper evaluates these paradigms' relative merits and trade-offs in reconstruction quality, rendering speed, and model size, illuminating the current landscape of dynamic 3DGS and addressing key challenges in dynamic scene reconstruction.

%% file: sec/2_related_work.tex
\section{Related Works}
\label{sec:related_work}

The field of dynamic 3DGS has evolved rapidly, with researchers developing diverse approaches to address the challenges of reconstructing scenes that change over time. In this section, we review the current state of the art, categorizing methods into two distinct paradigms: structure-guided methods and gaussian-centric methods.

\subsection{Structured-Guided}
Structure-guided methods leverage additional structured information such as neural networks, canonical spaces, or grid-based frameworks to model deformations in dynamic scenes. These approaches separate the representation of the scene's structure from its temporal evolution, using auxiliary models to predict how Gaussian primitives should transform temporally.

Deformable 3DGS \cite{yang2023deformable3dgs} pioneered the use of deformable 3D Gaussians for dynamic scenes by learning them in a canonical space and employing an implicit deformation field implemented as an MLP to model spatial-temporal deformation. This approach allows for complex deformations while maintaining a consistent scene structure in the canonical space. To mitigate overfitting caused by inaccurate poses, they introduce an annealing smooth training paradigm with linearly decaying Gaussian noise. GauFRe \cite{liang2024gaufregaussiandeformationfields} extends this by decoupling dynamic and static scene modeling, applying similar deformation techniques only to the dynamic elements.

Also building on the foundation of \cite{yang2023deformable3dgs}, GaGS \cite{lu2024gagaussian} introduced a method that voxelizes Gaussian distributions and utilizes sparse convolutions to extract geometry-aware features for deformation learning, enhancing the system's ability to capture fine-grained structural changes. Another approach, DynMF \cite{kratimenos2024dynmf}, models dynamics by factorizing scene motion into a compact set of learnable basis trajectories, arguing that complex movements can be efficiently represented by combining these bases. For a different take on efficient representation, SC-GS \cite{huang2023sc} employs a sparse set of control points—significantly fewer than the Gaussians—to drive deformation. These control points utilize radial basis functions and Linear Blend Skinning to interpolate transformations across the dense Gaussian field, and the system adaptively prunes or creates control points based on learned parameters.

4D-GS \cite{Wu_2024_CVPR} employs multi-scale HexPlane \cite{Cao2023HEXPLANE} as its foundational representation, encoding both temporal and spatial information through grid-based structures. This approach uses multi-head decoders to predict different Gaussians' attributes separately, enhancing the model's expressiveness. DeformGS \cite{deformgs} incorporates local rigidity and isometric loss constraints while designing regularization based on conservation of momentum to achieve smoother motion trajectories.

Structure-guided methods often benefit from incorporating various priors to address the under-constrained nature of dynamic reconstruction. Park et al. \cite{som2024} demonstrate this by consolidating supervisory signals from depth maps and long-range 2D tracks to achieve consistent dynamic representations. Similarly, Guo et al. \cite{guo2024motion} develop a flow-augmentation method that leverages correspondence between 3D Gaussian motion and pixel-level flow to improve reconstruction quality.

\subsection{Gaussian-Centric}
Gaussian-centric methods encode scene changes directly into the Gaussian primitives, often without relying on auxiliary structured representations. These methods adjust Gaussian parameters over time using continuous functions or frame-by-frame updates.

Dynamic 3DGS \cite{luiten2023dynamic} introduced one of the earliest approaches in this category, where Gaussians maintain consistent appearance properties (color, opacity, size) while their spatial properties (position, orientation) evolve temporally. Their online reconstruction process uses each frame to initialize the next, applying physical constraints like local rigidity, local rotational-similarity, and long-term local-isometry to ensure plausible Gaussian motion and rotation.


Continuous temporal modeling emerged as a key advancement: Katsumata et al. \cite{katsumata_compact_2024} incorporated Fourier approximations over time and optical flow supervision to model attribute changes of 3DGS, ensuring continuous changes without introducing excessive parameters. E-D3DGS \cite{bae2024ed3dgs} introduced per-Gaussian latent embeddings alongside temporal embeddings.

This continuous representation trend continued with STG \cite{li2023spacetime}, which utilized Radial Basis Functions for modeling opacity changes and parametric polynomials for position and rotation. Gaussian-Flow \cite{lin2024gaussian} combined time-domain polynomials with frequency-domain Fourier series to create a dual-representation model, complemented by adaptive timestep scaling and temporal smoothing that enhanced stability.

GauMesh \cite{xiao_2024_bridging} proposed a framework that integrated triangle meshes within a single differentiable pipeline. It leveraged tracked meshes for representing smooth surfaces and detailed textures, while using 3D Gaussians, deformed via a grid-based feature volume, to capture complex geometry, performing alpha-blending between the two representations for the final rendering.

Yang et al. \cite{yang2023gs4d} pioneered a technique treating time as an integral dimension, designing specialized 4D parameters. Duan et al. \cite{duan:2024:4drotorgs} developed a rotor-based 4D representation for complex space-time transformations. Addressing the significant storage demands associated with direct 4D Gaussian optimization, 4D Scaffold GS \cite{cho_4d_2024} extends 3D scaffolding into 4D space. It utilizes sparse 4D grid-aligned anchors, each holding a compressed feature vector, to model sets of neural 4D Gaussians representing local spatio-temporal regions.

%% file: sec/3_methods.tex
\section{Preliminary: 3D Gaussian Splatting}
\label{sec:preliminary}

3DGS \cite{kerbl3Dgaussians} represents scenes using explicit 3D Gaussian primitives, enabling efficient novel view synthesis. Each Gaussian is parameterized by: (1) center position $\boldsymbol{\mu} \in \mathbb{R}^3$, (2) covariance matrix $\Sigma \in \mathbb{R}^{3 \times 3}$ decomposed as $\Sigma = R\,S\,S^T\,R^T$ (where $R$ is rotation from quaternion $\mathbf{q}$ and $S$ is diagonal scaling), (3) view-dependent color via Spherical Harmonics, and (4) opacity $\alpha \in [0,1]$.

For rendering, 3D Gaussians are projected to 2D via transformation matrix $W$ with Jacobian $J$, yielding 2D covariance $\Sigma' = J\,W\,\Sigma\,W^T\,J^T$. The final pixel color is computed through alpha-blending: $C(p) = \sum_{i=1}^{N} \alpha_i c_i \prod_{j=1}^{i-1} (1 - \alpha_j)$. Parameters are optimized by minimizing $\mathcal{L} = (1 - \lambda)\,\mathcal{L}_1 + \lambda\,\mathcal{L}_{\text{D-SSIM}}$ between rendered and ground-truth images. Adaptive density control refines the representation through pruning low-opacity Gaussians and densification via cloning or splitting based on gradients.

\section{Methodological Analysis}
\label{sec:method}

Following the categorization in Section \ref{sec:introduction}, we examine the key methodological adaptations applied to the foundational 3DGS framework for dynamic scene reconstruction, focusing on the distinguishing characteristics of each paradigm.

\subsection{Structure-Guided}
Structure-guided methods maintain canonical 3D Gaussians and predict temporal transformations via auxiliary structures. These methods introduce specialized loss functions and regularization to ensure geometric plausibility: SC-GS \cite{huang2023sc} employs As-Rigid-As-Possible (ARAP) loss to encourage locally rigid transformations, while DynMF \cite{kratimenos2024dynmf} uses L1 sparsity regularization and rigid loss for coherent motion trajectories. Training typically involves jointly optimizing Gaussian parameters and deformation network weights, often with annealing schedules to stabilize convergence \cite{yang2023deformable3dgs}. Timestamps are embedded (via sinusoidal or learned encodings) and fed to deformation networks alongside spatial inputs. Some methods bypass neural networks, such as Gaussian-Flow \cite{lin2024gaussian}, by optimizing time-domain polynomials and Fourier components directly for faster convergence with fewer parameters.

\subsection{Gaussian-Centric}
Gaussian-centric methods encode temporal dynamics directly into primitives, extending 3D Gaussians to 4D or parameterizing attributes with continuous functions. This introduces new regularization requirements: 4D-Rotor GS \cite{duan:2024:4drotorgs} enforces 4D spatiotemporal consistency loss and entropy loss for sharper opacity, while STG \cite{li2023spacetime} uses temporal radial basis functions requiring additional smoothness constraints. Optimization involves 4D transformation matrices with time-dependent parameters, often compressed via functional representations \cite{katsumata_compact_2024, lin2024gaussian}. Training strategies include temporal slicing—batching samples from adjacent time intervals to leverage local continuity \cite{duan:2024:4drotorgs}—and temporal weighting mechanisms that focus gradients on relevant contributions \cite{yang2023gs4d}. Advanced systems employ time-conditioned multi-pass optimization, accumulating losses across temporal slices before gradient updates to ensure consistent 4D trajectories \cite{duan:2024:4drotorgs}.

%% file: sec/4_results.tex
\section{Experiments}
\label{sec:results}
We evaluate representative methods from both paradigms on the D-NeRF dataset \cite{pumarola2020d}, a widely-used benchmark comprising eight synthetic scenes with animated objects exhibiting large motions and non-rigid deformations. We selected methods with publicly available codebases from each category and established static 3DGS \cite{kerbl3Dgaussians} as a baseline. All experiments were conducted on a NVIDIA RTX 4090 GPU.

We report Peak Signal-to-Noise Ratio (PSNR) \cite{hore_image_2010}, Structural Similarity Index Measure (SSIM) \cite{wang_image_2004}, and Learned Perceptual Image Patch Similarity (LPIPS) \cite{zhang_unreasonable_2018} for reconstruction quality, alongside frames per second (FPS), training time, and model size for computational performance. 

\begin{table*}[htbp]
\caption{Reconstruction Quality and Computational Performance on D-NeRF \cite{pumarola2020d} dataset. \textbf{Bold} denotes best performance and \underline{underline} denotes second place.}
\vspace{-5mm}
\begin{center}
\small
\setlength{\tabcolsep}{4pt}
\begin{tabular}{l|ccc|ccc}
    \hline
    \textbf{Methods} & \textbf{PSNR $\uparrow$} & \textbf{SSIM $\uparrow$} & \textbf{LPIPS $\downarrow$} & \textbf{FPS $\uparrow$} & \textbf{Train (min) $\downarrow$} & \textbf{Size (MB) $\downarrow$} \\
    \hline
    \multicolumn{7}{c}{\textit{Baseline}} \\
    3DGS \cite{kerbl3Dgaussians} & 23.19 & 0.930 & 0.080 & 170 & 10 & \textbf{10} \\
    \hline
    \multicolumn{7}{c}{\textit{Structure-Guided}} \\
    SC-GS \cite{huang2023sc} & \textbf{43.31} & \textbf{0.997} & \textbf{0.007} & 123 & 17 & 22 \\
    Deformable 3DGS \cite{yang_deformable_2023} & \underline{39.51} & \underline{0.991} & 0.012 & 85 & 26 & \underline{14} \\
    GaGS \cite{lu2024gagaussian} & 37.36 & 0.988 & \underline{0.010} & 12 & 15 & 48 \\
    DynMF \cite{kratimenos2024dynmf} & 36.89 & 0.983 & 0.019 & 192 & 20 & 19 \\
    GauFre \cite{liang2024gaufregaussiandeformationfields} & 34.80 & 0.982 & 0.021 & 112 & 13 & 21 \\
    4DGS \cite{Wu_2024_CVPR} & 34.05 & 0.980 & 0.025 & 104 & 13 & 17 \\
    \hline
    \multicolumn{7}{c}{\textit{Gaussian-Centric}} \\
    STG \cite{li2023spacetime} & 34.27 & 0.986 & 0.032 & 276 & 23 & 41 \\
    4D Rotor GS \cite{duan:2024:4drotorgs} & 34.26 & 0.970 & 0.033 & \textbf{1257} & \textbf{8} & 36 \\
    RealTime4DGS \cite{yang_real-time_2024} & 34.09 & 0.979 & 0.021 & 231 & 10 & 1432 \\
    4D Scaffold GS \cite{cho_4d_2024} & 33.57 & 0.966 & 0.025 & \underline{293} & 34 & 24 \\
    Compact 3DGS \cite{katsumata_compact_2024} & 32.19 & 0.970 & 0.040 & 210 & \underline{9} & 159 \\
    Dynamic 3DGS \cite{luiten2023dynamic} & 27.69 & 0.951 & 0.057 & 74 & 16 & 3014 \\
    \hline
\end{tabular}
\label{tab:results}
\end{center}
\vspace{-8mm}
\end{table*}

\subsection{Reconstruction Quality}
All dynamic methods substantially outperform the static 3DGS baseline, confirming the necessity of temporal modeling. Structure-guided methods demonstrate superior reconstruction fidelity, with SC-GS \cite{huang2023sc} leading across all metrics, followed by Deformable 3DGS \cite{yang2023deformable3dgs} and GaGS \cite{lu2024gagaussian}. Gaussian-centric methods show wider variation, with top performers \cite{li2023spacetime, duan:2024:4drotorgs, yang2023gs4d} achieving competitive but generally lower quality than leading structure-guided approaches.

\subsection{Computational Cost}
Gaussian-centric methods achieve significantly higher rendering speeds, with 4D Rotor GS \cite{duan:2024:4drotorgs} reaching 1257 FPS and several others \cite{cho_4d_2024, li2023spacetime, yang2023gs4d} exceeding structure-guided methods. Structure-guided rendering speeds vary considerably (12-192 FPS). The most significant difference lies in model size: structure-guided methods maintain compact representations (14-48 MB), while gaussian-centric methods vary widely, with some compact \cite{cho_4d_2024} but others exceeding 1 GB \cite{yang2023gs4d, luiten2023dynamic}.

%% file: sec/5_discussion.tex
\section{Discussion}
\label{sec:Discussion}

A key finding is the consistent trend towards higher reconstruction fidelity among structure-guided methods, particularly evident in the monocular D-NeRF setup. Methods like SC-GS \cite{huang2023sc}, which achieved a leading PSNR of 43.31 dB, along with strong performers like Deformable 3DGS \cite{yang2023deformable3dgs} and GaGS \cite{lu2024gagaussian}, effectively leverage auxiliary structures. These include MLP-based deformation fields acting on canonical spaces \cite{yang2023deformable3dgs}, sparse control points driving deformation \cite{yang2023gs4d}, or sparse convolutions for geometry-aware feature extraction \cite{lu2024gagaussian}, to impose strong spatio-temporal priors. Incorporating such structured information and associated priors, such as ARAP losses or multi-scale geometric features, appears crucial for enhancing temporal consistency and reconstruction reliability, especially given the under-constrained nature of learning from limited dynamic observations. Furthermore, by separating static representation (canonical Gaussians) from dynamic transformation (learned fields or control points), this paradigm generally maintains relatively compact model storage requirements.

Conversely, the gaussian-centric paradigm frequently excels in computational efficiency, particularly rendering speed. Methods encoding temporal changes directly into Gaussian parameters, often using continuous functions like Fourier series \cite{katsumata_compact_2024}, polynomials \cite{lin2024gaussian}, or Radial Basis Functions \cite{li2023spacetime}, can bypass the need to query potentially large deformation networks at render time. The extension to true 4D primitives, as seen in RealTime4DGS \cite{yang2023gs4d}, 4DGS \cite{Wu_2024_CVPR}, and notably 4D-Rotor GS \cite{duan:2024:4drotorgs}, pushes rendering speeds to exceptional levels (Table \ref{tab:results}), significantly exceeding both static 3DGS and structure-guided methods. However, this paradigm shows considerable variability in storage demands. While techniques like 4D Scaffold GS \cite{cho_4d_2024} achieve compactness through anchored representations, others that explicitly store 4D parameters or require per-frame optimization like Dynamic 3DGS \cite{luiten2023dynamic} can lead to substantial model sizes, posing scalability challenges for long-duration videos or high resolution videos.

This comparison highlights a crucial trade-off in dynamic 3DGS: structure-guided methods often prioritize reconstruction fidelity and model compactness, potentially sacrificing some rendering speed, whereas gaussian-centric methods prioritize real-time rendering, sometimes at the expense of quality or storage efficiency. The choice between paradigms depends heavily on the specific application requirements: interactive AR/VR might favor the speed of gaussian-centric methods, while high-fidelity content creation might lean towards the quality offered by structure-guided approaches.

The importance of regularization is evident across both paradigms. The under-constrained nature of dynamic reconstruction necessitates techniques to prevent overfitting, temporal jitter, and implausible deformations. Our analysis highlights the effectiveness of various strategies, including ARAP losses \cite{yang2023gs4d}, sparsity constraints \cite{kratimenos2024dynmf}, local rigidity and smoothness priors \cite{luiten2023dynamic}, and spatio-temporal consistency or entropy losses for 4D representations \cite{duan:2024:4drotorgs}. These regularizers are critical for achieving stable training and temporally coherent results.




Building on these findings, several future directions emerge. Addressing large-scale unbounded dynamic scenes requires advancements in model scalability and adaptability to handle more intense motions and complex environmental changes, pushing beyond current benchmark limitations. This necessitates improving scene decomposition strategies, developing more efficient dynamic representations, exploring faster training techniques, and incorporating better geometric priors. Furthermore, research into novel data structures and primitives is crucial for better representing complex geometries, handling acute temporal events like appearances/disappearances, and optimizing resource usage. Exploring hybrid paradigms, perhaps using structure-guidance to regularize efficient gaussian-centric representations, could offer a compelling balance. Developing adaptive methods that select or switch representations based on scene complexity, performing thorough real-world validation on diverse captured data, integrating multi-modal priors like depth or flow, and designing more lightweight regularization techniques are all vital steps forward.

\section{Conclusion}
\label{sec:conclusion}
Dynamic 3DGS presents a vibrant research area characterized by a fundamental trade-off between reconstruction fidelity and computational efficiency. The structure-guided and gaussian-centric paradigms offer distinct solutions, each with unique advantages and challenges. Continued innovation in representations, regularization, and hybrid approaches will be essential to overcome current limitations and unlock the full potential of this technology for creating truly interactive and realistic dynamic digital worlds.
